\documentclass{article}
\usepackage{ijcai24}

\usepackage{times}
\usepackage{soul}
\usepackage{url}
\usepackage[hidelinks]{hyperref}
\usepackage[utf8]{inputenc}
\usepackage[small]{caption}
\usepackage{graphicx}
\usepackage{amsmath}
\usepackage{amsthm}
\usepackage{booktabs}
\usepackage{algorithm}
\usepackage{algorithmic}
\usepackage[switch]{lineno}

\usepackage{algorithm}
\usepackage{algorithmic}
\usepackage{mathtools}
\usepackage{amssymb}
\usepackage{booktabs}
\usepackage{bm}
\usepackage{newfloat}
\usepackage{listings}
\usepackage{makecell}
\usepackage{multirow}

\usepackage[marginal]{footmisc}

\usepackage{color}
\usepackage{hyperref}
\hypersetup{
    colorlinks=true,    
    urlcolor=cyan,
    citecolor=black,
}

\title{Diffusion Model is Secretly a Training-free Open Vocabulary Semantic Segmenter}

\author{
Jinglong Wang$^{1\dag}$\and
Xiawei Li$^{1\dag}$\and
Jing Zhang$^{1*}$\and
Qingyuan Xu$^1$ \\
Qin Zhou$^1$ \and
Qian Yu$^1$ \and
Lu Sheng$^1$ \and
Dong Xu$^2$ 
\\
\affiliations
$^1$Beihang University\\
$^2$The University of Hong Kong\\
\emails
wjlzy@buaa.edu.cn,
zy2121108@buaa.edu.cn,
zhang\_jing@buaa.edu.cn
}

\begin{document}

\maketitle

\footnote{*Corresponding author.}
\footnote{\dag These authors contributed equally.}

\begin{abstract}
The pre-trained text-image discriminative models, such as CLIP, has been explored for open-vocabulary semantic segmentation with unsatisfactory results due to the loss of crucial localization information and awareness of object shapes. Recently, there has been a growing interest in expanding the application of generative models from generation tasks to semantic segmentation. These approaches utilize generative models either for generating annotated data or extracting features to facilitate semantic segmentation. This typically involves generating a considerable amount of synthetic data or requiring additional mask annotations. 
To this end, we uncover the potential of generative text-to-image diffusion models (e.g., Stable Diffusion) as highly efficient open-vocabulary semantic segmenters, and introduce a novel training-free approach named DiffSegmenter. The insight is that to generate realistic objects that are semantically faithful to the input text, both the complete object shapes and the corresponding semantics are implicitly learned by diffusion models. We discover that the object shapes are characterized by the self-attention maps while the semantics are indicated through the cross-attention maps produced by the denoising U-Net, forming the basis of our segmentation results.
Additionally, we carefully design effective textual prompts and a category filtering mechanism to further enhance the segmentation results. Extensive experiments on three benchmark datasets show that the proposed DiffSegmenter achieves impressive results for open-vocabulary semantic segmentation. The project page is available at \href{https://vcg-team.github.io/DiffSegmenter-webpage/}{https://vcg-team.github.io/DiffSegmenter-webpage/}.

\end{abstract}

\section{Introduction}
\begin{figure}[t]
\centering
\includegraphics[width=\hsize]{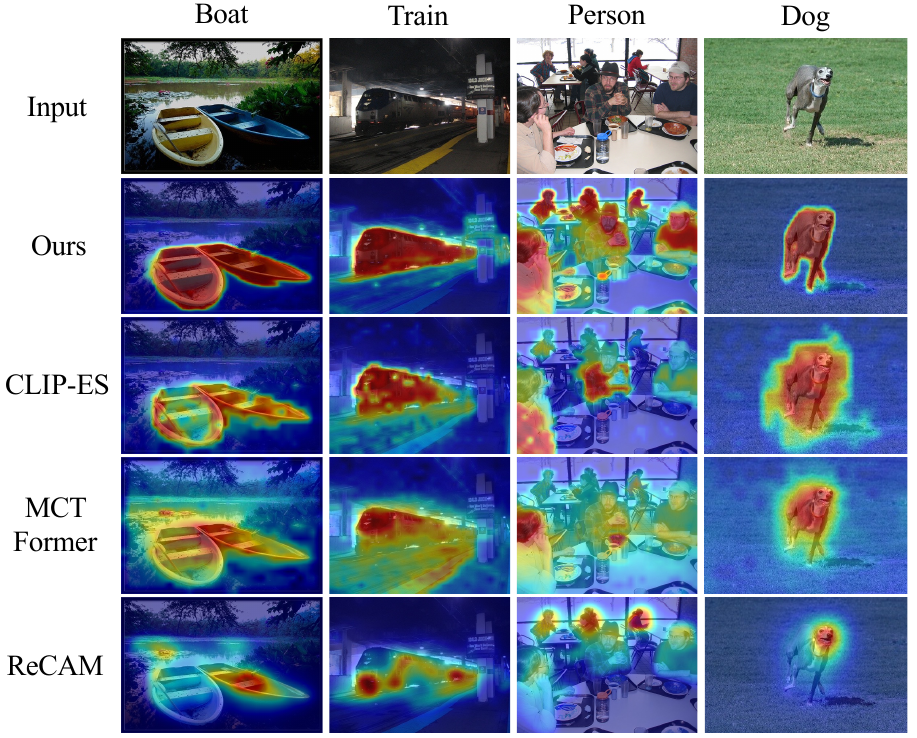} 
\caption{Segmentation score maps generated by our proposed DiffSegmenter and previous discriminative methods.}
\label{figure:1}
\vspace{-1.5em}
\end{figure}
The process of gathering and annotating images with pixel-level labels is labor-intensive and time-consuming. Models trained solely on fully annotated data are restricted to specific categories, which limits their scalability. As a result, there is an increasing focus on developing open vocabulary semantic segmentation methods. These approaches are considered more practical alternatives to fully supervised approaches, as they reduce the reliance on extensive annotation and enable segmentation across a wider range of categories.

Large-scale image-text pre-trained models like CLIP~\cite{radford2021learning} have showcased remarkable zero-shot transferability across diverse downstream tasks. As a result, researchers have started to investigate the potential of leveraging these models for open-vocabulary semantic segmentation in zero-shot~\cite{ding2022decoupling,liang2023open}
setting or weakly supervised~\cite{xie_2022_clims,Lin_2023_clipes} setting.
However, the alignment process based on contrastive learning by the discriminative image-text pre-trained models may unintentionally result in the loss of crucial localization information and awareness of object shapes, since only the most discriminative visual features are learnt through text-image contrastive learning.

Recently, there has been a growing interest in exploring the application of generative models, such as diffusion models, beyond image generation tasks, particularly in the field of semantic segmentation
~\cite{wu2023diffumask,karazija_2023_ovd,baranchuk2021label,xu2023open}. 
While these methods show promise, they still require specific training or complex data synthesis procedures. For instance, SegDiff and MedSegDiff rely on annotated data for supervised segmentation. For open-vocabulary segmentation tasks, ODISE and SegDiff require additional mask annotations from a large dataset for training. DiffuMask and OVDiff involve complex data synthesis process.

In this paper, we delve deeper into the capabilities of generative text-to-image diffusion models for semantic segmentation.
In contrast to discriminative image-text pre-trained models that model the posterior distribution $p(\bm{c}|\bm{x})$ of a whole image, diffusion models estimate the conditional probability density $p(\bm{x}|\bm{c})$ for generating semantically meaningful objects.
To generate realistic objects that are semantically faithful to the input text, both the complete object shapes and the corresponding semantics are implicitly learned.
By using Bayes’ theorem with a proper prior $p(\bm{c})$, it becomes straightforward to convert $p(\bm{x}|\bm{c})$ into pixel-level posterior distribution $p(\bm{c}|\bm{x})$ for semantic segmentation.
Moreover, large-scale diffusion models trained by huge amount of image-text pairs allow them to encode rich visual-semantic correspondence priors, further potentially enabling semantic understanding within whole vocabulary of the pre-trained diffusion models. By exploiting the per-pixel discriminative power of text-to-image diffusion models, it not only benefits semantic segmentation tasks, but also has high impact on some downstream tasks, such as controllable image editing.

To this end, we uncover the potential of generative text-to-image diffusion models as highly efficient training-free semantic segmenters, and introduce a novel training-free approach named DiffSegmenter. The proposed DiffSegmenter enables open-vocabulary semantic segmentation using off-the-shelf diffusion models without the need for any additional learnable modules or any parameter tuning. 
We discover that the object shapes are characterized by the self-attention maps while the semantics are indicated through the cross-attention maps produced by the denoising U-Net. The combination of both directly results in the segmentation score maps.
Additionally, we carefully design effective textual prompts and a category filtering mechanism by harnessing the power of the pre-trained BLIP~\cite{li_2022_blip} model to further enhance the segmentation results. 

Figure~\ref{figure:1} shows a comparison of the segmentation score maps obtained from our proposed DiffSegmenter and previous discriminative model-based methods, showing the improved segmentation mask. 
Extensive experiments on PASCAL VOC 2012~\cite{everingham_2010_pascalvoc}, MS COCO 2014~\cite{lin_2014_microsoftcoco} datasets, and Pascal Context~\cite{mottaghi_2014_pascalvoccontext} verify that our generative approach to semantic segmentation, leveraging the inherent capabilities of diffusion models, achieves impressive results for open-vocabulary semantic segmentation in both zero-shot setting and image-level weakly supervised setting. Moreover, our method shows a strong potential to assist the downstream image editing.

In summary, our contributions are three-fold:
\begin{itemize}
\setlength\itemsep{0em}
    \item 
    We uncover the potential of generative text-to-image conditional diffusion models as highly efficient training-free semantic segmenters by modeling  $p(\bm{x}|\bm{c})$.
    \item We propose a novel training-free open-vocabulary semantic segmentation method, DiffSegmenter, which fully exploits the attention layers in the denoising U-Net of diffusion models and carefully designs the textual prompts for semantic enhancement.
    \item Extensive experiments show that the proposed DiffSegmenter achieves impressive results for semantic segmentation and has high potential on assisting image editing. 
\end{itemize}

\section{Related Work}
\subsection{CLIP-based Open-vocabulary Segmentation}
Large-scale image-text pre-trained models like CLIP~\cite{radford2021learning} have shown promising zero-shot classification capabilities. It has recently been extended to dense prediction tasks such as semantic segmentation. CLIP models are discriminative models trained by aligning images with global category label, denoted as $p(\bm{c}|\bm{x})$, for an input image $\bm{x}$. However, semantic segmentation tasks necessitate per-pixel predictions denoted as $p(\bm{c}|\bm{x}_{h,w})$, where $\bm{c}$ is the class label for each individual pixel in the image $\bm{x}_{h,w}$.

Hence, existing methods tackle the adaptation of CLIP models for semantic segmentation through three primary approaches. 1) A large scale pixel-wise annotated data or pseudo labels are used to finetune the CLIP image encoder~\cite{rao2022denseclip,zhou_2022_maskclip}.
2) Additional pre-trained mask generators~\cite{liang2023open,ghiasi2022scaling} super-pixel grouping methods~\cite{xu_2023_ovs,ding2022decoupling,luddecke2022image} are required to obtain the object mask, while CLIP is employed solely as a proposal/group classifier. 3) Since discriminative models only activate the most discriminative parts of class objects, complex multi-round refinement mechanisms are employed to enhance the class activation maps obtained from CLIP models as segmentation results~\cite{xie_2022_clims,Lin_2023_clipes}.

However, the alignment process based on contrastive learning employed by the discriminative image-text pre-trained models may unintentionally result in the loss of crucial localization information and awareness of object shapes, leading to unsatisfactory segmentation results.

\begin{figure*}[h]
\centering
\includegraphics[width=1\textwidth]{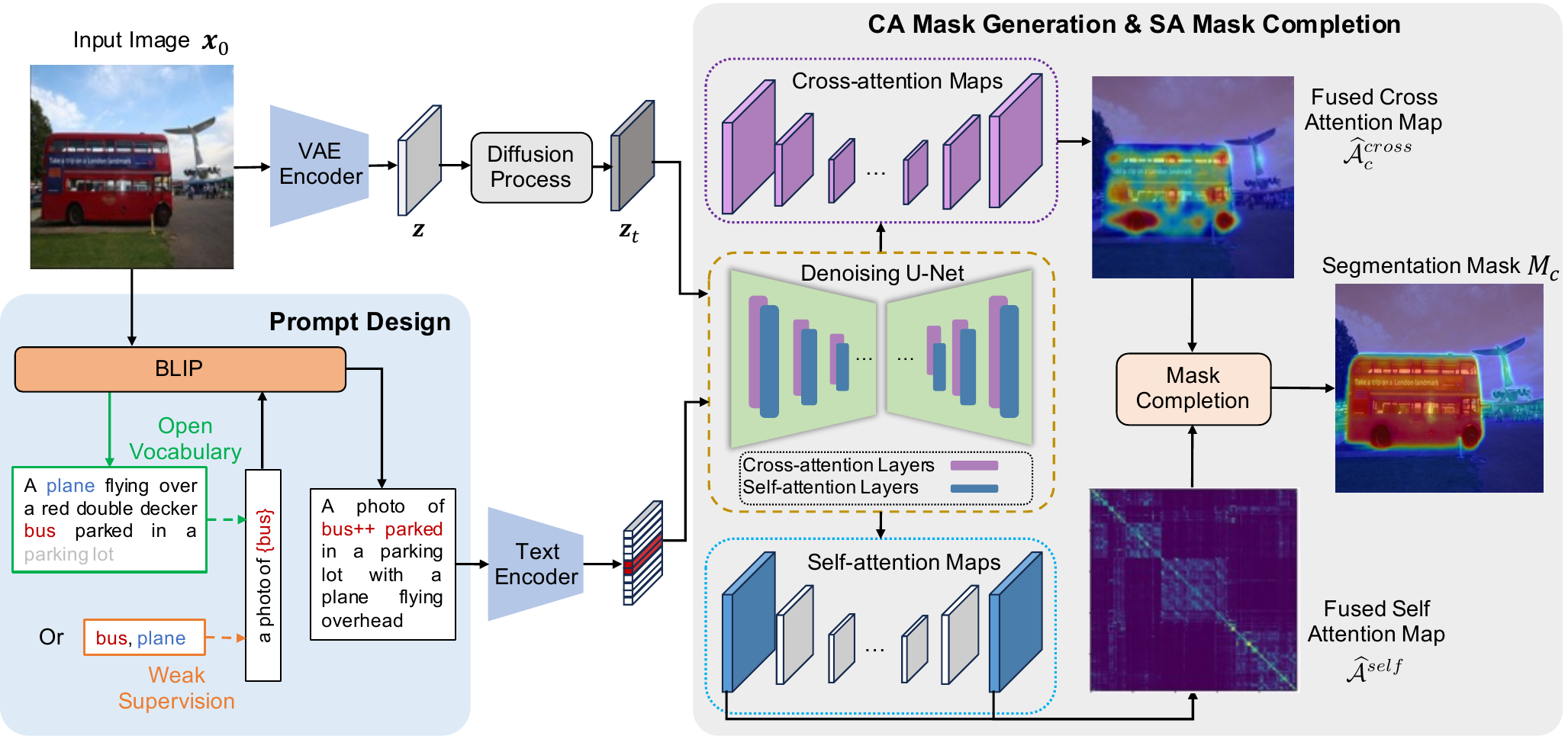} 
\caption{\textbf{Overview of the proposed DiffSegmenter.} An input image and enhanced candidate class tokens by the BLIP-based prompt design module are fed into an off-the-shelf pre-trained conditional latent diffusion model.
The fused cross-attention maps produced by the denoising U-Net are treated as the initial segmentation score maps, which is further refined and completed by the fused self-attention maps of the U-Net. Note that the parameters of all the involved models are frozen without any tuning.}
\label{figure:2}
\vspace{-1em}
\end{figure*}

\subsection{Diffusion Models for Perception Tasks}

Recently, diffusion models have undergone significant developments and emerged as prominent generative models in contemporary research~\cite{ho2020denoising,rombach2022high}. 
The cross-attention mechanism that allows for conditional synthesis in diffusion model is also exploited for different tasks by previous work, such as image editing~\cite{DBLP:conf/iclr/HertzMTAPC23}, sketch synthesis~\cite{xing_2023_diffsketcher}, and interpreting the Stable Diffusion model~\cite{tang_2022_daam}. 

Nevertheless, diffusion models possess not only powerful generative capabilities but also can be cleverly utilized in various perception tasks. The Diffusion Classifier demonstrated that density estimation derived from diffusion models, which generate images from text, can be employed for zero-shot classification tasks without requiring additional training~\cite{li2023your}. Leveraging the energy-based modeling nature of diffusion models, they can be applied to unsupervised compositional concept discovery, facilitating effective representation of semantic information in images~\cite{liu2023unsupervised}. TF-ICON can leverage off-the-shelf diffusion models to perform cross-domain image-guided composition without requiring additional training or finetuning~\cite{lu2023tficon}.

Diffusion models have also been exploited for semantic segmentation tasks through various approaches. Directly inputting the latent variables from diffusion models into segmentation networks enables semantic segmentation even with limited training samples~\cite{baranchuk2021label}. DiffuMask leverages diffusion models to generate images and pixel-level image annotations, thereby training a high-performing semantic segmentation model~\cite{wu2023diffumask}. However, both of these approaches are not open-vocabulary semantic segmentation solutions. ODISE leverages large-scale text-to-image diffusion models and discriminative models to construct a panoramic segmentation scheme~\cite{xu2023open}. OVDiff initially utilizes diffusion models to generate prototypes for multiple classes and then matches pixel features with these prototypes during actual segmentation, thereby determining the class of each pixel~\cite{karazija_2023_ovd}. Although these two approaches yield excellent results, they still require model training and the use of a pretrained segmentation network for actual segmentation. In contrast to previous works, our approach does not need addtional training and does not rely on a pretrained segmentation network.
\vspace{-1em}
\section{Methodology}
\subsection{Diffusion Model for Segmentation: A Baseline}
A generative model that models $p(\bm{x}|\bm{c})$ can be easily adapted for semantic segmentation via Bayes' theorem.
The conditional diffusion models such as Stable Diffusion~\cite{rombach2022high} define the conditional probability of $\bm{x}_0$ as,
\begin{equation}
    p_\theta(\bm{x}_0|\bm{c})=\int_{\bm{x}_{1:T}}p(\bm{x}_T)\prod_{t=1}^Tp_{\theta}(\bm{x}_{t-1}|\bm{x}_t,\bm{c})d\bm{x}_{1:T},
\end{equation}
where $p(\bm{x}_T)$ is fixed to $\mathcal{N}(0,I)$. Diffusion models are trained to minimize the variational lower bound (ELBO)~\cite{blei_2017_variational} of the log-likelihood,
\begin{equation}
    \log p_{\theta}(\bm{x}_0|\bm{c})\leq \mathbb{E}_q \left[ \log \frac{p_{\theta}(\bm{x}_{0:T},\bm{c})}{q(\bm{x}_{1:T}|\bm{x}_0)} \right].
\end{equation}
Inspired by diffusion classifier~\cite{li2023your}, the ELBO is,
\begin{equation}
\label{eq:ELBOloss}
    -\mathbb{E}_{t,\epsilon}\left[\|\epsilon-\epsilon_\theta(\bm{x}_t,\bm{c}) \|^2 \right].
\end{equation}
As the loss for the ELBO defined in Eq.~\ref{eq:ELBOloss} is calculated on a per-pixel basis, we can leverage Bayes' theorem 
to obtain pixel-level classification results. The baseline method demonstrates reasonable results by carefully set the diffusion time steps and utilizing a large number of samples for Monte Carlo estimation. However, it is worth noting that this approach requires a substantial computational cost due to the high number of samples required. 

Furthermore, it is important to note that the diffusion process has the potential to compromise the structural information of the original image. And thus the fine-grained details and local pixel-level semantics may become less distinguishable or even blurred during the diffusion progresses, which is detrimental to the semantic segmentation results. 

By delving deeper into the capabilities of conditional latent diffusion models and their ability to model conditional distributions $p(\bm{x}|\bm{c})$ through the utilization of cross-attention modules, this paper introduces a new attention-based method, named DiffSegmenter. 
DiffSegmenter excels not only in semantic segmentation tasks but also exhibits versatility for a wider range of downstream tasks, including but not limited to image editing. By utilizing DiffSegmenter, more accurate and refined image segmentation results can be obtained, thereby providing more effective editing control.
\subsection{Proposed DiffSegmenter}
\paragraph{Motivation.}
With the conditional denoising autoencoder $\epsilon_\theta(\bm{x}_t,t,\bm{c})$, Stable Diffusion utilizes cross-attention mechanism to establish the relationship between text features and latent vsiual features. This allows us to control the denoising process based on the text condition for generating different images.
In other words, Stable Diffusion can model conditional distributions of the form $p(\bm{x}|\bm{c})$ thanks to the cross-attention mechanism.

In the case of segmentation tasks, if we assume a uniform prior over $\{\bm{c}_i\}$, the formulation of Bayes' theorem
simplifies to $p(\bm{c}_i|\bm{x})=\frac{p(\bm{x}|\bm{c}_i)}{\sum_j p(\bm{x}|\bm{c}_j)}$. It can be concluded that $p(\bm{c}_i|\bm{x})\propto p(\bm{x}|\bm{c}_i)$. Since Stable Diffusion can approximate the distribution $p(\bm{x}|\bm{c}_i)$ with ELBO, it can also be used for discriminative tasks. 
Building upon this insight, we utilize cross-attention from the diffusion model to establish the relationship between the distributions $p(\bm{c}_i|\bm{x})$ and $p(\bm{x}|\bm{c}_i)$.

The cross-attention map calculates the similarity between the query and key matrices (Q and K) and applies the weighted sum to the value matrix (V). This means that the spatial features are augmented with the most relevant textual features for each pixel. A higher similarity leads to larger activation values in $\mathcal{A}$, indicating a closer relationship between the current pixel and the corresponding text. Therefore, we simply use the cross-attention maps calculated by the pre-trained conditional latent diffusion models as the mask basis for semantic segmentation, without the need for any additional learnable modules or any parameter tuning.

\paragraph{Overview.}
An overview of our method is shown in Figure~\ref{figure:2}. We feed an input image and candidate classes into an off-the-shelf pre-trained conditional latent diffusion model.
By extracting the cross-attention maps produced by the denoising U-Net, we obtain the initial segmentation score maps. We further propose to refine and complete the semantic score maps with the self-attention maps of the U-Net, which effectively capture pairwise pixel affinities to link pixels with similar features. Moreover, we carefully design effective textual prompts and category filtering mechanism based on BLIP model to further enhance the results.

\subsection{Cross-attention-based Score Map Generation}
We feed an input image to be segmented to the VAE encoder $\mathcal{E}$ to obtain a latent variable $\bm{z} = \mathcal{E}(\bm{x}_0)$. In each time-step $t$, we add Gaussian noise to $\bm{z}$, resulting in $\bm{z}_t = \sqrt{\bar{\alpha_t}}\bm{z} + \sqrt{1-\bar{\alpha_t}}\epsilon$, where $\epsilon \sim N(0, I)$. 
The deep spatial features are represented as $\varphi(\bm{z}_t) \in \mathbb{R}^{H\times W \times C}$, where $H$ is the height of the feature map, $W$ is the width, and $C$ is the number of channels. These features are linearly mapped to the query matrix $Q = \ell_Q(\varphi(\bm{z}_t))$. The text prompt $\mathcal{P}$ is encoded by the text encoder and mapped to $\tau_{\theta}(\mathcal{P})\in \mathbb{R}^{N\times D}$, where $N$ is the length of the text tokens and $D$ is the latent projection dimension. The text is further linearly mapped to the key matrix $K = \ell_K(\tau_{\theta}(\mathcal{P}))$ and the value matrix $V = \ell_V(\tau_{\theta}(\mathcal{P}))$.

The cross-attention maps are computed as:
\begin{equation}
    \mathcal{A}^{cross} = \text{Softmax}(\frac{QK^T}{\sqrt{d}}),
\end{equation}
where $ \mathcal{A}^{cross} \in \mathbb{R}^{H\times W \times N\text{(reshaped)}} $, and $\mathcal{A}^{cross}_c\in {\mathbb{R}^{H\times W}}$ represents the cross-attention for a specific class token $\bm{c}$. The output of the cross-attention $\widehat{\varphi}(\bm{z}_t) = \mathcal{A}^{cross}V$
is used to update the spatial features $\varphi(\bm{z}_t)$.

\begin{figure}[!h]
\vspace{-1em}
\centering
\includegraphics[width=1\linewidth]{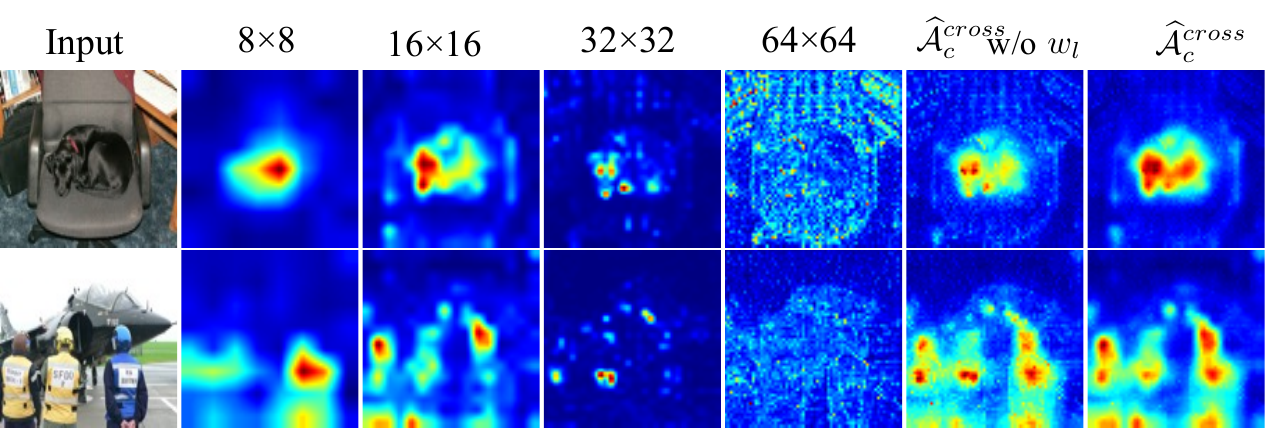} 
\vspace{-1em}
\caption{Cross-attention maps of different layers.}
\label{figure:3}
\vspace{-1.5em}
\end{figure}

We observed that different cross-attention layers have complementary abilities to capture semantic information as shown in Figure~\ref{figure:3}. The attention maps from the smaller feature maps can precisely localize the objects-of-interest while the attention maps from the larger feature maps capture more fine-grained object details.Therefore, we fuse the cross-attention maps of class $\bm{c}$ from multiple layers with different importance weights $w_l$, which obtains:
\begin{equation}
    \widehat{\mathcal{A}}_c^{cross} = \sum_{l \in L} w_l\cdot \mathcal{A}_{c,l}^{cross} \in \mathbb{R}^{H\times W},
\end{equation}
where $\mathcal{A}_{c,l}^{cross}$ is the cross-attention map of class token $\bm{c}$ of the $l$-th layer in the UNet, and $\sum_{l \in L} w_l=1$ is the importance weights.Attention maps from four different layers (e.g., $8\times8$, $16\times16$, $32\times32$, $64\times64$) of the UNet are used and interpolated to the same size for fusion in our method. The fused cross-attention map serves as the initial segmentation score maps in our method.

\subsection{Self-attention-based Score Map Completion}

Although the score maps generated by cross-attention maps could successfully localize the objects of interest, it has been observed that the produced score maps often lack clear object boundaries and may exhibit internal holes. 
It is worth noting that there exists attention between latent representations in UNet, namely the self-attention map $\mathcal{A}_{l}^{self}\in \mathbb{R}^{HW\times HW}$. The Q, K, and V (i.e., Query, Key, and Value) in self-attention are all based on latent visual features, enabling the establishment of correlations between different pixels. This endows self-attention with the ability to perform region completion, which can compensate for the incomplete activation regions in cross-attention.By considering different time steps and layers, we obtain the fused self-attention map, 
\begin{equation}
    \widehat{\mathcal{A}}^{self} = \frac{1}{L}\sum_{l \in L}{\mathcal{A}_l^{self}}\in \mathbb{R}^{HW\times HW},
\end{equation}
In our method, instead of fusing self-attention maps from all the layers of the UNet, we specifically select and fuse the attention maps of the encoder and decoder with a latent feature size of $64\times64$, where more fine-grained pixel-level affinity weights are captured.

To address the issues of incomplete segmentation score maps generated by cross-attention maps, we introduce a mask refinement mechanism by multiplying the cross-attention maps with pixel affinity weights obtained from self-attention maps. The refined segmentation score maps for class $\bm{c}$ is,
\begin{equation}
    M_c=\text{norm}(\widehat{\mathcal{A}}^{self}\cdot vec(\widehat{\mathcal{A}}_c^{cross})),
\end{equation}
where $\text{norm}(\cdot)$ is min-max normalization to ensure the segmentation score maps are appropriately scaled,$vec(\widehat{\mathcal{A}}_c^{cross})\in \mathbb{R}^{HW \times 1}$, and $vec(\cdot)$ is a vectorization operation of a matrix. Therefore, the final segmentation score maps $M_c$ are simply produced during the denoising inference process of the pre-trained text-conditional diffusion models, without any additional additional learning modules or any parameter tuning.

\subsection{Prompt Design for Semantic Enhancement}

\paragraph{Augmented Prompts.}
The quality of the attention-based segmentation score maps is also highly influenced by the textual inputs. Intuitively, providing more comprehensive and detailed descriptions of the objects-of-interest can lead to better localization. Building on this intuition, 
in addition to the class token, we also incorporate corresponding adverbs and adjectives that provide more detailed attributes of the objects. The cross-attention maps of the class names and the adverbs or adjectives are fused to obtain the segmentation score maps.

\paragraph{Class Token Re-weighting.}
To put greater emphasis on the class token corresponding to the object-of-interest, we introduce a re-weighting mechanism for the class token embedding. By assigning a higher weight to the class token, we highlight the target objects while suppress the background for better segmentation results. The re-weighted class token is denoted as ``[class]++''.

\subsection{Open-vocabulary Semantic Segmentation}
In zero-shot open-vocabulary semantic segmentation, we generate $C$ candidate score maps by individually feeding each candidate class into the text encoder and compute the $M_c$ for the input image. The final segmentation results are obtained by taking the maximum value of each pixel by comparing $C$ score maps. To reduce the computational cost, we pre-filter the candidate categories using two strategies. Firstly, we feed the input image to the BLIP model, which generates captions describing the image content. From these captions, we extract the nouns, which represent the main objects or entities present in the image. These extracted nouns then serve as the filtered categories, narrowing down the candidate categories for semantic segmentation. Secondly, we feed the input image and all the candidate classes into CLIP model and select the classes with cosine similarity larger than 0.97. The union of the selected classes by BLIP and CLIP models is used as our final filtered categories for a given image.

In weakly-supervised open-vocabulary semantic segmentation, the final segmentation results of the training data are produced from the prompted ground-truth object labels of the input image. And any off-the-shelf segmentation models can be used for training the segmentation models.

\section{Experiments}
\subsection{Datesets and Implementation Details}
To validate our methodology, we conducted experiments on tasks involving open-vocabulary semantic segmentation in both weakly-supervised setting with image-level class supervisions and completely zero-shot setting.

For the task of weakly supervised semantic segmentation, we evaluated our approach on the PASCAL VOC 2012~\cite{everingham_2010_pascalvoc}. PASCAL VOC 2012 consists of 21 categories (including one background class). The augmented set comprises 10,582 images for training  and 1449 images for testing.
For zero-shot open-vocabulary semantic segmentation, we evaluate our approach on PASCAL VOC 2012(VOC), Pascal Context(Context), and COCO-Object(Object) datasets. Pascal Context~\cite{mottaghi_2014_pascalvoccontext} have 60 classes(including backgroud class). Since our method is entirely train-free, we solely validate our approach on the validation set. The VOC, Context, and Object datasets comprise 1449, 5105, and 5000 images respectively.

We employed the BLIP~\cite{li_2022_blip} model to generate textual descriptions for input images. For zero-shot open-vocabulary segmentation, we extracted nouns from the generated captions and compared them to all the candidate categories to compute similarity. We utilize the outcomes of CLIP~\cite{Lin_2023_clipes} classification to complement the extraction of nouns. This process yielded labels for the images, which were subsequently incorporated as prompts into the BLIP model to obtain comprehensive descriptions of the image.
We validated our approach using Stable Diffusion v1.5 with frozen pre-trained parameters.

For the weakly supervised semantic segmentation task, we followed the approach ~\cite{Lin_2023_clipes,xie_2022_clims} to assess the quality of the initial Class Activation Maps (CAMs). After applying DenseCRF~\cite{krahenbuhl_2011_densecrf} to obtain pseudo-labels, we proceeded to train a fully supervised segmentation network, such as DeepLabV2. 
We employ the mean Intersection over Union (mIoU) as the evaluation metric for all experiments.

\subsection{Results on Open-vocabulary Segmentation}
Table~\ref{open1} shows a comparison of our method to previous work on zero-shot open-vocabulary semantic segmentation. We evaluate mIoU on three datasets: VOC, Context, and Object. Compared to previous training-based approaches, our method exhibits a significant improvement in mIoU across different datasets, showing the strong zero-shot generalization capability of our method. Our method only performs poorer than a concurrent work OVDiff on VOC and Context datasets. We argue that OVDiff necessitates a complex image synthesis process and involve additional pre-trained segmenters and feature extractors for prototype generation. By contrast, our method is simple and efficient.
\begin{table}[h]
\renewcommand{\arraystretch}{1}
\centering
\resizebox{\linewidth}{!}{
\begin{tabular}{c|ccc}
\toprule[1.5pt]
Method&VOC&Context&Object\\
\hline
\multicolumn{2}{l}{\textit{Training-involved}} \\
ReCo~\cite{shin_2022_reco}&25.1&19.9&15.7\\
ViL-Seg~\cite{liu_2022_vilseg}&37.3&18.9&-\\
MaskCLIP~\cite{zhou_2022_maskclip}&38.8&23.6&20.6\\
TCL~\cite{cha_2023_tcl}&51.2&24.3&30.4\\
CLIPpy~\cite{ranasinghe_2022_clippy}&52.2&-&32.0\\
GroupViT~\cite{xu_2022_groupvit}&52.3&22.4&-\\
ViewCo~\cite{ren_2023_viewco}&52.4&23.0&23.5\\
SegCLIP~\cite{luo_2023_segclip}&52.6&24.7&26.5\\
OVSegmentor~\cite{xu_2023_ovs}&53.8&20.4&25.1\\\hline
\multicolumn{2}{l}{\textit{Training-free}} \\
OVDiff~\cite{karazija_2023_ovd}&\textbf{67.1}&\textbf{30.1}&\underline{34.8}\\ 
DiffSegmenter (Ours) &\underline{60.1} & \underline{27.5} & \textbf{37.9} \\ 
\bottomrule[1.5pt]
\end{tabular}}
\caption{Results of zero-shot open-vocabulary semantic segmentation on three benchmark datasets.}
\label{open1}
\end{table}

\begin{figure}[h]
\centering
\setlength{\abovecaptionskip}{2em}
\includegraphics[width=0.95\linewidth]{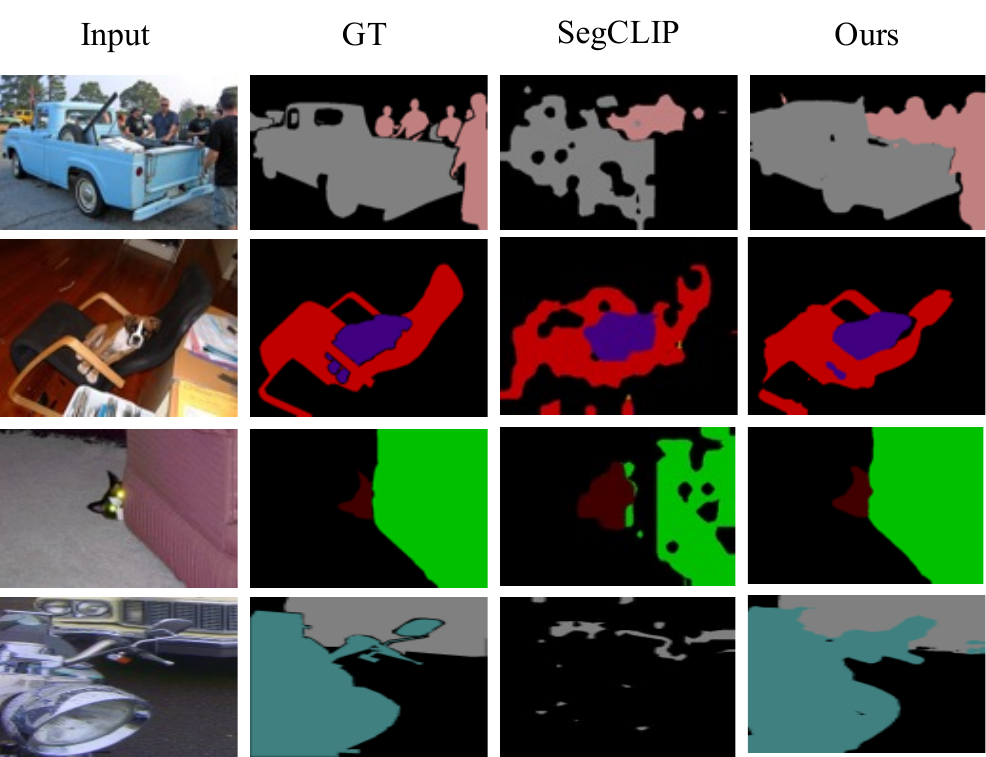} 
\vspace{-1em}
\caption{Qualitative results of DiffSegmenter and SegCLIP baseline for zero-shot open-vocabulary segmentation. }
\label{figure:open1}
\vspace{-1em}
\end{figure}

Figure~\ref{figure:open1} illustrates the qualitative segmentation results of a comparison between DiffSegmenter and the SegCLIP baseline. The results indicate that our approach achieves a more complete segmentation results of the objects of interest, with clearer mask boundaries. More results can be found in the supplementary material.

\subsection{Results on Weakly-supervised Segmentation}
Table~\ref{weak1} shows the segmentation results of the training set of VOC with image-level class labels. Our method results in an mIoU of 70.5\%, which significantly outperforms most of the discriminative model-based baselines with image-level supervision. Our method is also comparable to the state-of-the-art CLIP-ES model, which is a very strong baseline. However, CLIP-ES rely on both image-level supervision and language supervision while our method only rely on image-level supervision. Moreover, CLIP-ES uses a complex synonym fusion strategy to enrich the category names, which may also be used together with our method to further improve the results.
In Figure~\ref{figure:1}, we showcase segmentation score maps for some example images, illustrating that our method generates more complete semantic score maps than baseline methods.

\begin{table}[h!]
\renewcommand{\arraystretch}{1.0}
\centering
\resizebox{0.8\linewidth}{!}{
\begin{tabular}{c|c}
\toprule[1.5pt]
\textbf{Method} & \textbf{VOC train} \\
\hline
\multicolumn{2}{l}{\textit{Image-level Supervsion}} \\
IRN~\cite{ahn_2019_irn}&48.8\\
SC-CAM~\cite{chang_2020_sccam}&50.9\\
SEAM~\cite{wang_2020_seam}&55.4\\
AdvCAM~\cite{lee_2021_advcam}&55.6\\
RIB~\cite{lee_2021_rib}&56.5\\
OoD~\cite{lee_2022_ood}&59.1\\
MCTfomer~\cite{xu_2022_mct}&61.7\\
DiffSegmenter (Ours) &\textbf{70.5}\\ \hline
\multicolumn{2}{l}{\textit{Image-level Supervision+Language Supervision}} \\
CLIMS~\cite{xie_2022_clims}&56.6\\
CLIP-ES~\cite{Lin_2023_clipes}&70.8\\
\bottomrule[1.5pt]
\end{tabular}}
\caption{Segmentation results of on PASCAL VOC 2012 train sets with image-level object labels. }
\vspace{-0.5em}
\label{weak1}
\end{table}

To further evaluate the quality of the generated segmentation masks, we utilized the segmentation masks in the training set as pseudo-labels for training a segmentation network. By following the state-of-the art methods, such as CLIMS and CLIP-ES, we employed the DeepLabV2 architecture pretrained on ImageNet-1k
as our segmentation network. The results are presented in Table~\ref{weak2}, where the segmentation model trained with the masks generated by our approach achieved 69.1\% and 68.6\% on the validation and test sets, which is comparable to the state-of-the-art methods.

\begin{table}[t]
\renewcommand{\arraystretch}{1.2}
\centering

\resizebox{0.8\hsize}{!}{
\begin{tabular}{cc|cc}
\toprule[1.5pt]
Method&Backbone&Val&Test\\
\hline
\multicolumn{4}{l}{\textit{Image-level Supervsion}} \\
AdvCAM~\cite{lee_2021_advcam}&R101&68.1&68.0\\
RIB~\cite{lee_2021_rib}&R101&68.3&\textbf{69.1}\\
ReCAM~\cite{chen_2022_recam}&R101&68.5&68.4\\
DiffSegmenter (Ours)&R101&\textbf{69.1}&\textbf{68.6}\\
\hline
\multicolumn{4}{l}{\textit{Image-level Supervision+Language Supervision}} \\
CLIMS~\cite{xie_2022_clims}&R101&69.3&68.7\\
CLIP-ES~\cite{Lin_2023_clipes}&R101&71.1&71.4\\
\bottomrule[1.5pt]
\end{tabular}}

\caption{Weakly-supervised semantic segmentation results on PASCAL VOC 2012 validation and test sets.}
\label{weak2}
\end{table}

\subsection{Ablation Studies and Analyses}
We present the results of ablation study to our method in Table~\ref{table_ablation} to evaluate the effectiveness of each component.

\begin{table}[h!]
\renewcommand{\arraystretch}{1}
\centering
\resizebox{0.8\linewidth}{!}{
\begin{tabular}{ccccc|c}
\toprule[1.5pt]
\multicolumn{5}{c|}{\textbf{Method}} & VOC train\\
 $\widehat{\mathcal{A}}^{cross}_c$ & $\widehat{\mathcal{A}}_{all}^{self}$ & $\widehat{\mathcal{A}}^{self}$ & BLIP & ``++'' & \textbf{mIoU}\\
 \hline
 w/o $w_l$ &&&\checkmark & \checkmark & 61.25\\ 
 w/o $w_l$ &\checkmark &&\checkmark & \checkmark & 65.01\\ 
 w/o $w_l$ &&\checkmark &\checkmark & \checkmark & 67.89\\ 
 \hline
\checkmark & & \checkmark &  &  & 65.32\\ 
\checkmark & & \checkmark & \checkmark &  & 67.99\\ 
\checkmark & & \checkmark &  & \checkmark & 69.46\\ 
\hline
\checkmark & & \checkmark &\checkmark & \checkmark & \textbf{70.49}\\ 
\bottomrule[1.5pt]
\end{tabular}
}
\caption{Ablations study.}
\label{table_ablation}
\vspace{-1em}
\end{table}

\begin{figure}[h!]
\vspace{-0.5em}
\centering
\includegraphics[width=\linewidth]{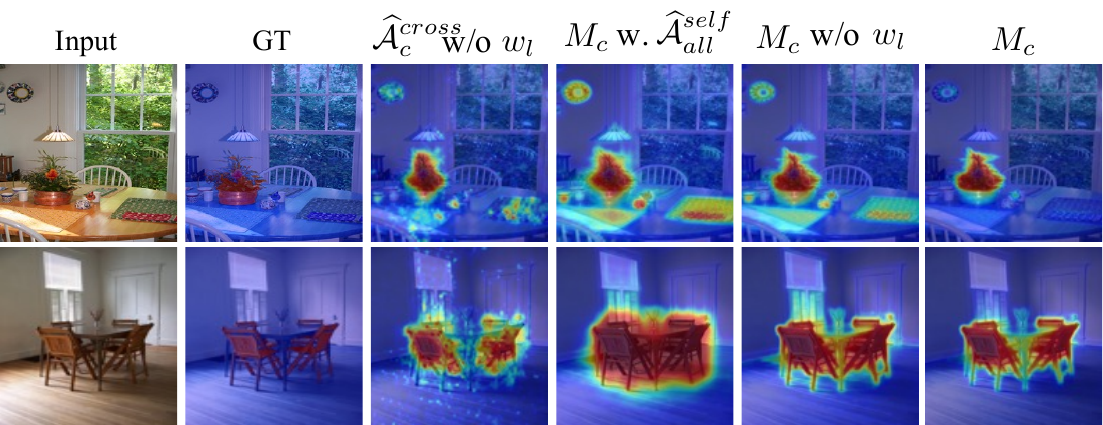} 
\caption{Comparison of different mask generation variants.}
\label{figure:ablation1}
\vspace{-1em}
\end{figure}

\paragraph{Effect of different mask generation strategies.}
Table~\ref{table_ablation} presents the mIoU evaluation results obtained on the PASCAL VOC 2012 training set when using different mask generation strategies proposed in our method. The visualized comparison results are shown in Figure~\ref{figure:ablation1}. 
When using only cross-attention and averaging the values of the four cross-attention layers (i.e., $\widehat{\mathcal{A}}_{c}^{cross}$ w/o $w_l$), we can obtain an mIoU of 61.25\%, which is already comparable to the state-of-the-art methods with image-level supervision. The incorporation of self-attention in all layers of the model (i.e., $\widehat{\mathcal{A}}_{all}^{self}$), the obtained $M_c$ with $\widehat{\mathcal{A}}_{all}^{self}$ leads to a significant improvement in performance. This verifies that self-attention provides rich image information that complements the limitations of cross-attention and enhances the activation maps. Furthermore, focusing solely on the highest resolution self-attention $\widehat{\mathcal{A}}^{self}$, the obtained $M_c$ leads to additional advancements, resulting in more complete image boundaries.

As shown in Figure~\ref{figure:3}, we observe that different cross-attention layers contain varying degrees of semantic information. For example, the strongest semantic information is captured by the cross-attention at sizes of $8\times8$ and $16\times16$. Instead of simple averaging, we applied weighted sum to the cross-attention from different layers. We empirically used weights of [0.3, 0.5, 0.1, 0.1] for the cross-attention at sizes of $8\times8, 16\times16, 32\times32$, and $64\times64$, respectively. By reinforcing the semantic information, the final results are further improved compared to the averaging operation.

\begin{figure}[h!]
\centering
\vspace{-1em}
\includegraphics[width=1\linewidth]{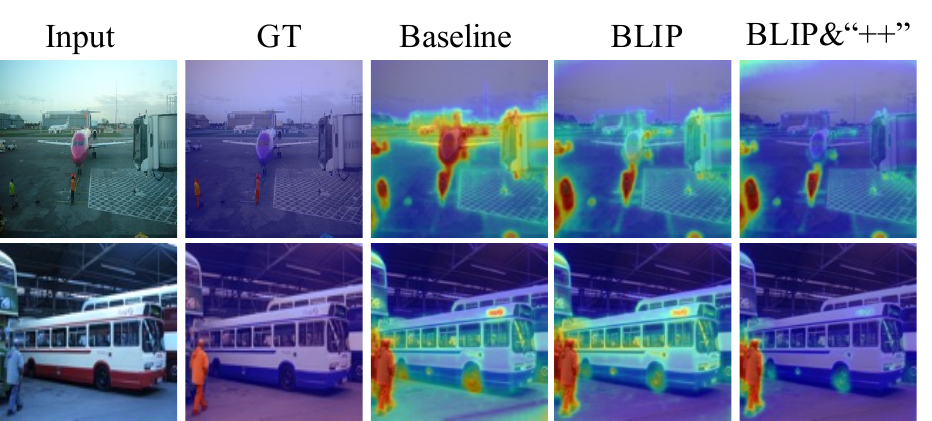} 
\caption{Comparisons of different prompt designs.}
\label{figure:ablation2}
\vspace{-1em}
\end{figure}

\paragraph{Evaluations of the Prompt Design.}
Table~\ref{table_ablation} also provides the related ablation study for prompt design while fixing the attention map variant. Introducing BLIP allows us to generate conditional captions, providing additional textual descriptions for target categories. Attention maps corresponding to modified adjectives or adverbs that describe the target category can help in segmenting the target object. By changing the re-weighting the target category words, we can make the attention map focus more on the target category while suppressing the activation of the background or other objects. This strategy greatly enhances the segmentation results. In Figure~\ref{figure:ablation2}, we can have consistent observations. ``BLIP'' refers to using BLIP to complete the prompt and generate a conditional caption. The meaning of ``++'' is to apply the class token reweighting method specifically to target category word.

\paragraph{Different Timesteps.}
In our method, generating attention maps only requires one denoising timestep using the diffusion model. To investigate the impact of the denoising timesteps, we conducted more experiments specifically focusing on the timestep. The experimental results are shown in Table~\ref{table_ablation3}. It can be observed that the segmentation performance is the best at t=100. If the timestep is too large or too small, the segmentation results drop to some extent. By averaging the results from multiple sampling timesteps, the best results can be achieved. Thus, in weakly supervised setting, we employ average of multiple sampling timesteps, while in zero-shot setting, we simply use t=100 to accelerate the inference speed.

\begin{table}[h]
\renewcommand{\arraystretch}{1.2}
\centering
\resizebox{0.8\linewidth}{!}{
\begin{tabular}{c|c|c|c|c|c}
\toprule[1.5pt]
\textbf{Method} & t=1& t=50& t=100& t=150& \textbf{Avg.}\\
\hline 
\textbf{mIoU} & 69.10 & 69.94 & 70.30 & 69.69 & 70.49\\
\bottomrule[1.5pt]
\end{tabular}
}
\caption{Results of different timesteps. \textbf{Avg.} is calculated by averaging the results of t=1,t=50,t=100 and t=150.}
\label{table_ablation3}
\end{table}
\vspace{-0.5em}
\subsection{Impacts on Image Editing}
\begin{figure}[h!]
\vspace{-1em}
\centering
\includegraphics[width=1\linewidth]{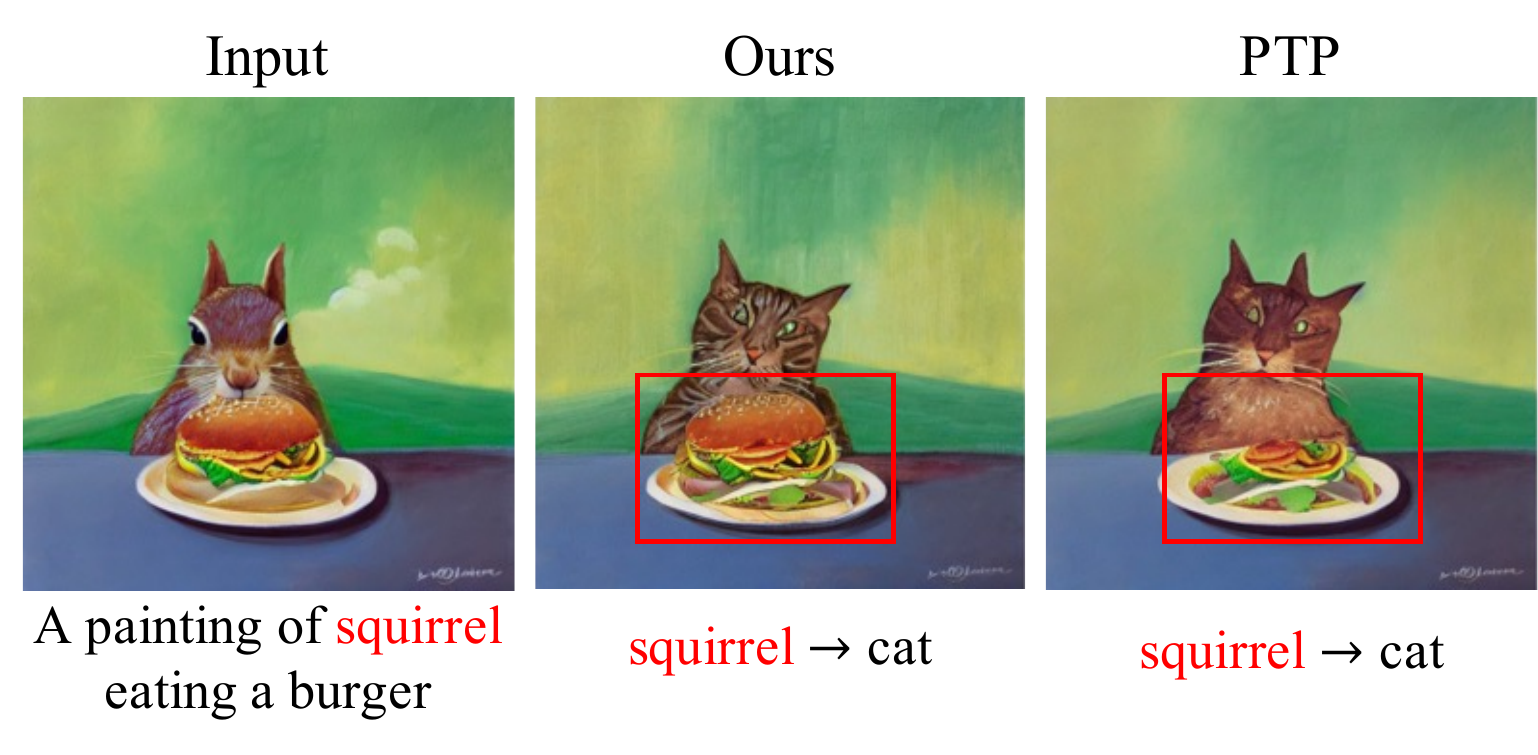} 
\vspace{-1.8em}
\caption{Impacts on image editing compared with PTP.}
\label{figure:comparision_ptp}
\vspace{-1em}
\end{figure}

Our method can also be adapted for applications in downstream tasks that rely on precise semantic segmentation masks, such as image editing. As shown in Figure~\ref{figure:comparision_ptp}, by using our method to obtain more complete masks for attention replacement, we can better preserve the background region and only modify the corresponding foreground region. This demonstrates the superiority of our approach over Prompt-to-Prompt (PTP)~\cite{DBLP:conf/iclr/HertzMTAPC23} which directly replaces cross-attention, as it enables us to achieve more accurate and controlled editing results.

\vspace{-0.5em}
\section{Conclusion}
This paper presents an innovative technique called DiffSegmenter, which allows for open-vocabulary semantic segmentation using readily available diffusion models, without requiring additional learnable modules or parameter tunings. The proposed methodology maximizes the capabilities of attention layers within the denoising U-Net of diffusion models. By harnessing the power of the pre-trained BLIP model, carefully designed textual prompts are incorporated to enhance semantic quality.
One limitation of our method is that the Stable Diffusion model relies on the latent features of the input image, which may result in the disappearance of small objects in the small feature maps of the latent space. This issue might be potentially addressed by employing other text-to-image diffusion models that are based on the original images or have larger latent feature map sizes.

\bibliographystyle{named}
\bibliography{ijcai24}

\end{document}